\documentclass{article} 
\usepackage{iclr2019_conference,times}

\usepackage{amsmath,amsfonts,bm}

\def\eqref#1{equation~\ref{#1}}

\def\1{\bm{1}}

\DeclareMathAlphabet{\mathsfit}{\encodingdefault}{\sfdefault}{m}{sl}
\SetMathAlphabet{\mathsfit}{bold}{\encodingdefault}{\sfdefault}{bx}{n}

\usepackage{hyperref}
\usepackage{url}
\usepackage{graphicx}

\usepackage[caption=false,font=normalsize]{subfig}
\usepackage{xspace}
\usepackage[ruled]{algorithm2e}
\usepackage{multirow, makecell}
\usepackage{enumitem}

\title{Better Accuracy with Quantified Privacy: \\Representations Learned via Reconstructive Adversarial Network}

\author{Sicong Liu, 
Xidian University,
\texttt{scliu007@gmail.com} \\
\textbf{Anshumali Shrivastava, 
Rice University, 
\texttt{anshumali@rice.edu}} \\
\textbf{Junzhao Du,  
Xidian University, 
\texttt{dujz@xidian.edu.cn}} \\
\textbf{Lin Zhong,
Rice University, 
\texttt{lzhong@rice.edu} }\\
}

\newcommand{\ie}{\emph{i.e.},\xspace}
\newcommand{\eg}{\emph{e.g.}\xspace}

\newcommand{\systemname}{{\sf RAN}\xspace}
\newcommand{\systemnameposs}{{\sf RAN's}\xspace}

\newcommand\rev[1]{\textcolor{black}{#1}}

\begin{document}

\maketitle

\begin{abstract}

The remarkable success of machine learning, especially deep learning, has produced a variety of cloud-based services for mobile users. Such services require an end user to send data to the service provider, which presents a serious challenge to end-user privacy. To address this concern, prior works either add noise to the data or send features extracted from the raw data.  They struggle to balance between the utility and privacy because added noise reduces utility and raw data can be reconstructed from extracted features.


This work represents a methodical departure from prior works: we balance between a measure of privacy and another of utility by leveraging adversarial learning to find a sweeter tradeoff. We design an encoder that optimizes against the reconstruction error (a measure of privacy), adversarially by a Decoder, and the inference accuracy (a measure of utility) by a Classifier. The result is \systemname, a novel deep model with a new training algorithm that automatically extracts features for classification that are both private and useful.  

It turns out that adversarially forcing the extracted features to only conveys the intended information required by classification leads to an implicit regularization leading to better classification accuracy than the original model which completely ignores privacy. Thus, we achieve better privacy with better utility, a surprising possibility in machine learning! We conducted extensive experiments on five popular datasets over four training schemes, and demonstrate the superiority of \systemname compared with existing alternatives.

\end{abstract}

\section{introduction}
\label{sec:introduction}
Today's most robust and accurate models are boosted by deep learning techniques, which benefit a lot of mobile intelligent services, such as speech-based assistant (\eg Siri), face recognition enabled phone-unlock (\eg FaceID).
However, the uncontrolled submission of raw sound, image, and human activity data from mobile users to \textit{service provider} has well-known privacy risks~\cite{bib:abadi2016:abadi}.
\rev{
For example, the underlying correlation detection, re-identification and other malicious mining~\cite{bib:dwork2017:ARSA, bhatia2016privacy}.
Different from pinning hopes on service providers to anonymise data for privacy-preserving, we present to encode each piece of raw data in the end-user side and only send the encoded data to the service provider. And the encoded data must be both \textit{private} and \textit{useful}.
\textit{Privacy} can be quantified by the risk of sensitive raw data disclosure given the encoded data.
For classification services, \textit{utility} can be quantified by the inference accuracy, achieved by the service provider using a discriminative model.
}

%

Existing solutions addressing the privacy concern struggle to balance between above two seemingly conflicting objectives: privacy vs. utility.
An obvious and widely practiced solution to the above problem is to transform the raw data into features and upload the features only, like Google Now~\cite{url:googlenow}; 
Google Cloud Machine Learning Engine also provides API to preprocess the raw data into engineering features before uploading~\cite{url:googlecloud}.
This solution not only alleviates the privacy concern but also reduces the mobile data usage. 
However, it does not provide any quantifiable privacy guarantee.
It is well known that we can reconstruct the raw data from the features~\cite{bib:mahendran2015:CVPR}.
As a result, \cite{bib:ossia17:arxiv} further apply dimensionality reduction and add noise to the features before sending them to the service provider, which unfortunately result in inference accuracy degradation.

Unlike previous work, we aim to systematically derive deep features for a sweeter tradeoff between \textit{privacy} and \textit{utility} using deep neural networks, by leveraging adversarial training.
Our key idea is to judiciously combine generative learning, for maximizing reconstruction error,
and discriminative learning, for 
minimizing discriminative error.
Specifically, we present Reconstructive Adversarial Network (RAN), an end-to-end deep model with a new training algorithm.
RAN controls two types of descent gradients, i.e., reconstruction error and discriminative error, in back-propagation process to guide the training of a feature extractor or Encoder.
%

%
%
%
%
\rev{
Defining the exact adversarial attacker and finding the right measurement for privacy is an open problem in itself~\cite{bib:mendes2017:access}.
In this paper, we quantify \textit{Privacy} using an intuitive metric, \ie the difficulty of reconstructing raw data via a generative model, or the \emph{reconstruction error}. In this case, the adversarial attacker is defined as a data reconstructor.
}
Therefore, as shown in Figure~\ref{fig_ran_design}, a \systemname consists of three parts: a feature extractor (Encoder), a utility discriminator (Classifier), and an adversarial reconstructor (Decoder).
The output of the Encoder feeds to the input of the Classifier and the Decoder.
We envision the Encoder runs in mobile devices and processes raw data into features. The Classifier runs in an untrusted platform, e.g. the cloud. A malicious party can seek to reconstruct the raw data from the features using the Decoder.
There is no theoretic guarantee on end-to-end training the colloborated discriminative model and generative model. Therefore, we present a novel algorithm to train the RAN via an adversarial process, \ie training the Encoder with a Classifier first to improve intermediate features' utility for discriminative tasks and confronting the Encoder with an adversary Decoder to enhance the features' privacy.
%
All three parts,  Encoder, Classifier and Decoder, are iteratively trained using gradient descent.
%
From the manifold perspective, the two separate flows across \systemnameposs Encoder, Decoder and Classifier, \ie decent gradients of discrimination error and reconstruction error from the end of Classifier and Decoder in back-propagation, guide the exact model parameter updating, which can iteratively derive the privacy-specific and utility-imposed feature manifold.

Using MNIST~\cite{data:mnist1998:LeCun}, CIFAR-10\cite{data:cifar}, ImageNet~\cite{data:imagenet}, Ubisound~\cite{bib:sicong2017:IMWUT} and Har~\cite{data:Har} benchmark datasets, we show \systemname is effective in training an Encoder for end users to generate deep features that are both private and useful. 

Surprisingly, we observe adversarial learned features to remove redundant information, for privacy, even surpass the accuracy of the original model. Removing redundant information enhances the generalization. See $\S$~\ref{sec:experiment} and $\S$ A for more details. This better generalization is as auspicious illustration that in practice, with machine learning, we can gain both utility and privacy at the same time.

In the rest of this paper, we elaborate \systemnameposs design in $\S$ 2 and evaluate the performance of \systemname in $\S$ 3. 
We next review the related work in $\S$ 4 and conclude this work in $\S$ 5. 
We finally present the theoretic interpretation of \systemname in $\S$~\ref{sec:discuss}. 

\section{Design of RAN}
\label{sec:design}
This section first formulates the privacy preserving problem, and then elaborates on \systemnameposs design.
%

\subsection{Problem Definition of Mobile Data Privacy Preserving}\label{sec:problem}
\begin{figure}[t]
\centering
\includegraphics[width=.78\textwidth]{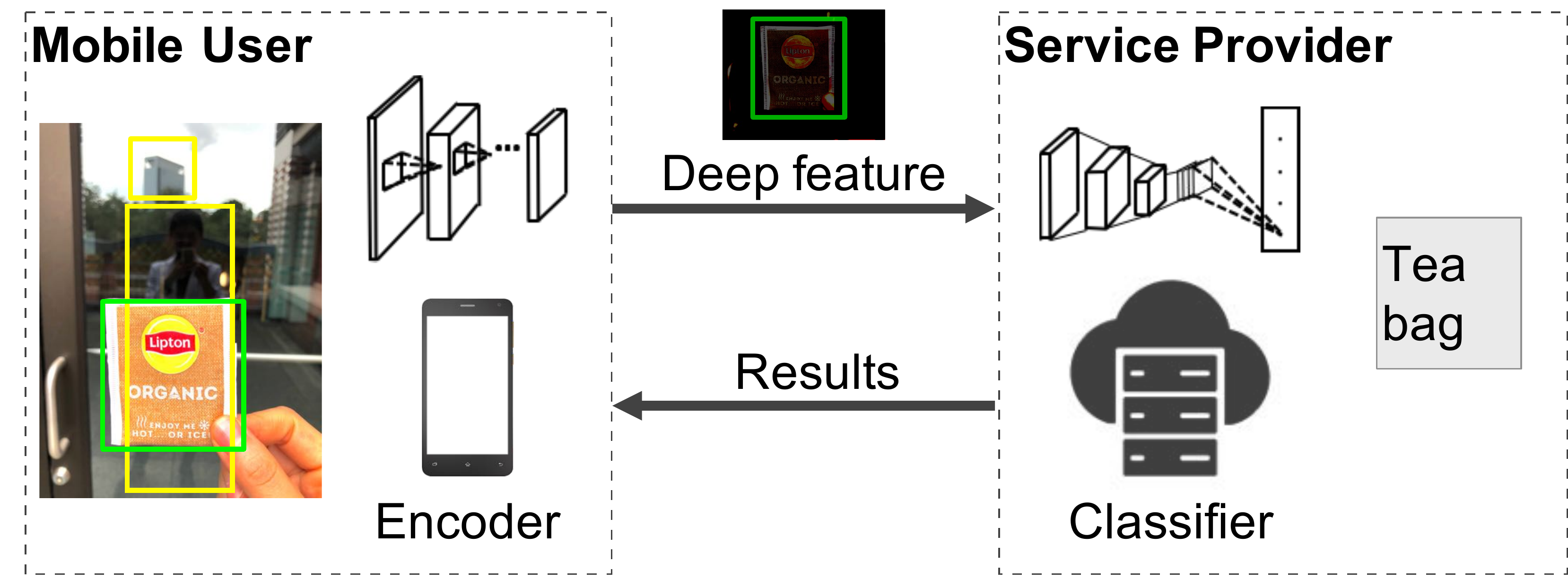}
\caption{Framework of mobile data privacy preserving. Mobile users leverage the learned Encoder to generate deep features from the raw data (\ie "tea bag" picture) before submit it. And the service provider use the learned Classifier based on the received deep features, to recognize the object in the picture and recommend a seller.}
\label{fig_privacy_program}
\end{figure}

Many services exist today to analyze data from end users.
In this work, we do not trust service providers for the privacy
of data: they could be malicious or subject to malicious exploits.
For example, as shown in Fig \ref{fig_privacy_program}, an end user takes a picture of a product and send it to a cloud-based service to find a place to purchase it, which is indeed a service Amazon provides.
A lot of sensitive information could accidentally come with the picture, such as personal information and user location in the background.
%

Our key insight is that most services actually do not need the raw data. Therefore, the mobile user can encode raw data into features through a multi-layer Encoder (E) on the client side and only deliver features to the service provider.
Such features ideally should have following two properties:
%
\textbf{Utility:} contain enough essential information of raw data so that they are useful for the intended service, e.g., high accuracy for object recognition;
%
\textbf{Privacy:} it is hard to recover the original information of raw data based on perturbed features through a reverse deep model~\cite{zhang2016privacy}.
%

\subsection{Utility and Privacy Metrics}
In this work, we focus on classification services. Therefore, utility is quantified as the inference accuracy of a discriminative model, employed by the service provider. 
\rev{Since defining the exact adversarial attacker and finding the right measurement for privacy is an open problem in itself~\cite{bib:mendes2017:access},
this paper quantifies privacy by an intuitive metric, \ie the reconstruction error in a reversed deep model, $X$, employed by a malicious party.
The reconstruction error measures the risk of original data disclosure.}
Since the Encoder is distributed to mobile users, we assume it is available to both service providers and potential attackers. That is, both the service provider and the malicious party can train their models using raw data and their corresponding Encoder output.
As such we can restate the desirable properties for the Encoder output as: 
\begin{equation}
\label{equ_DRL_R2}
\begin{split}
\text{\textbf{Utility}:} \quad & \underset{E}{Max} \  prob(Y_i'= Y_i), i \in \mathbf{T}\\
\text{\textbf{Privacy}:} \quad & \underset{E}{Max} \ \underset{X}{Min} \  |I_i-I_i'|^2,i \in \mathbf{T} \\
\end{split}
\end{equation}
where, $prob(Y_i'=Y_i)$ denotes the correct inference probability, \ie accuracy,  in the classification service with the testing data $\mathbf{T}$. $Y_i'$ and $Y_i$ is the inference class and the true label, respectively.
$|I_i-I_i'|^2$ is the  Euclidean distance, \ie reconstruction error, between the raw data $I_i$ and the mimic data $I_i'$ reconstructed by a malicious party with the Encoder output.

The first objective (Utility) is well-understood for discriminative learning.
It can be achieved via a standard optimization process, \ie minimizing the cross entropy between the predicted label $Y_i'$ and ground truth $Y_i$ in a supervised manner~\cite{bib:kruse2013:CI}. The inner part of the second objective, $\underset{X}{Min}\ |I_i-I_i'|^2$, is also well-understood for generative learning. On the other hand, the outer part $\underset{E}{Max}\ |I_i-I_i'|^2$ is the opposite, \ie maximizing the reconstruction error.
Therefore, the Encoder and the reverse deep model employed by the malicious party ($X$) are adversarial to each other in their optimization objectives.

Achieving above two objectives at the same time is challenging,
since utility, \ie maximized accuracy, and privacy, \ie maximized reconstruction error,
are conflicting objectives to the feature extractor, \ie Encoder.
When improving Utility, the Encoder must extract features to represent the relevant essence of data; when improving Privacy, the Encoder can discard the utility-relevant essence of the data. If not done properly, the Encoder output optimized for Utility leads to effective data reconstruction by a reverse model and therefore poor Privacy~\cite{bib:rifai2011:ICML}.


\subsection{Architecture of RAN}\label{sec:ran}
\begin{figure}[t]
\centering
\includegraphics[width=.88\textwidth]{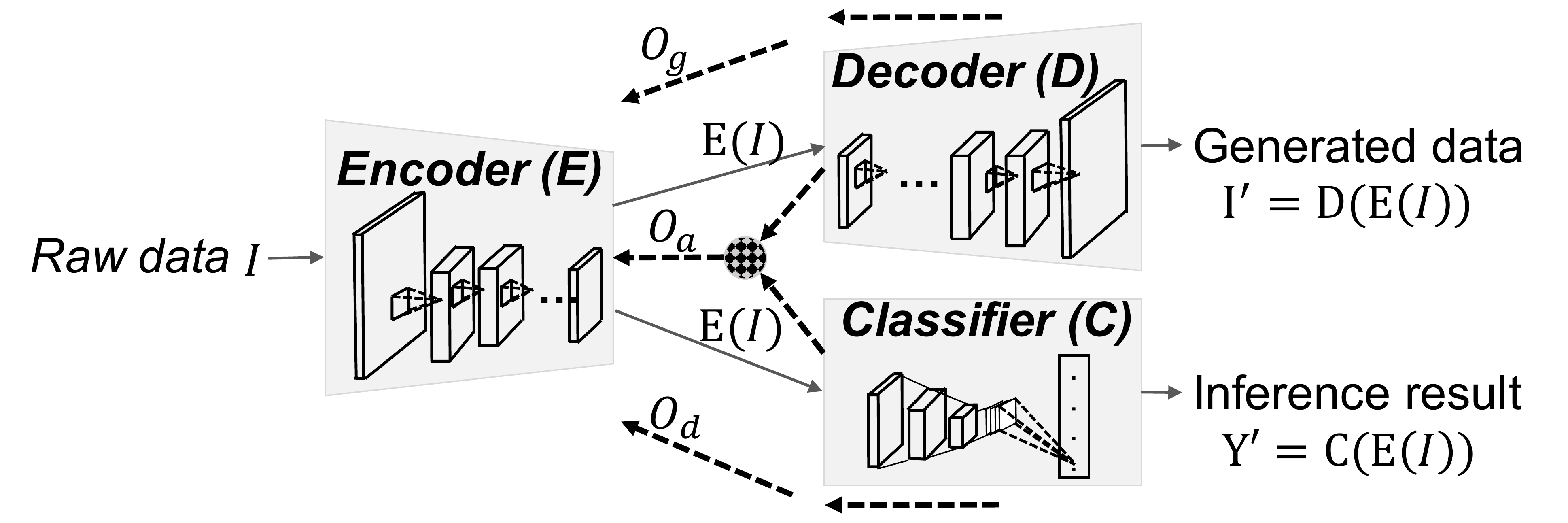}
\caption{Architecture of reconstructive adversarial network (RAN).}
\label{fig_ran_design}
\end{figure}

To tackle above challenges, we present \systemname to train a feature extractor, \ie Encoder, with good trade-offs between privacy and utility.
As shown in Fig \ref{fig_ran_design}, \systemname employ two additional neural network modules, Decoder ($D$) and Classifier ($C$), to train the Encoder ($E$). The Classifier simulates the intended classification service; when RAN is trained by the service provider, the Classifier can be the same discriminative model eventually used. The Decoder simulates a malicious attacker that attempts to reconstruct the raw data from the Encoder output.
All the three modules are end-to-end trained to establish the Encoder (E) for end-users to extract deep features $E(I)$ from raw data $I$.
%
%
The training is an iterative process that will be elaborated in \S \ref{sec:algorithm}. Below we first introduce \systemnameposs neural network architecture, along with some empirically gained design insights. 

\begin{itemize}[leftmargin=*]
\item The \textbf{Encoder (E)} consists of an input layer, multiple convolutional layers, pooling layers, and batch-normalization layers. 
%
%
We note that the clever usage of pooling layers and batch-normalization layers contribute to deep feature's utility and privacy.
%
The batch-normalization layer 
helps the features' utility because it normalize the activation to avoid being too high or too low thus has an regularization affect~\cite{bib:ioffe2015:arxiv}.
It contributes to features' privacy as well since it is hard for Decoder to recover detail information from normalized features.
And then, the max-pooling layer 
is helpful to enhance feature's privacy, because none of un-pooling techniques can recover fine details from size-reduced features through shifting small parts to precisely arrange them into a larger meaningful structure~\cite{bib:milletari2016:3DV}.

\item The \textbf{Decoder (D)} is a usual Encoder turned upside down, composed of multiple un-pooling layers~\cite{bib:mahendran2015:CVPR} and deconvolutional layers~\cite{bib:zeiler2010:deconvolutional}.
%
%
%
We note that the use of Decoder in training Encoder is to simulate a malicious party. After obtaining a (binary) version of the Encoder, a malicious party is free to explore any neural architectures to reconstruct the raw data. 
\rev{In this paper, we choose a worst possible Decoder, \ie an exactly layer-to-layer reversed architecture to mirror the Encoder. That is, we assume a powerful adversarial Decoder that knows the Encoder's operations and connections in training. 
}
We also note that the architecture and training algorithm of RAN can easily incorporate other architectures as the Decoder.

\item the \textbf{Classifier (C)} builds a multi-layer perceptron (MLP) to process deep features and output inference results with several full-connected layers~\cite{bib:kruse2013:CI}.
As we noted for the Decoder above, a service provider can explore any neural architectures for its discriminative model, given the Encoder. The reason we choose this specific architecture to train the Encoder is
because some of the most successful CNN architectures, \eg VGG and AlexNet, which can be viewed as as the Encoder plus the Classifier of our choice.
\end{itemize}

\subsection{Training Algorithm of RAN}\label{sec:algorithm}

\begin{algorithm}[t]
 \LinesNumbered
 \KwIn{Dataset $\mathbf{T}$}
 \KwOut{RAN's Weights $\{\theta_e, \theta_d, \theta_c \}$ }
 Initialize $\theta_e$, $\theta_d$, $\theta_c$ \;
 \For{$n$ epochs}{
Sample mini-batch $I$ of $m$ samples from $\mathbf{T}$\;
\For{$k$ steps}{
Update $\theta_e$ and $\theta_c$ by gradient ascent with learning rate $l_1$: minimize $O_d $ \;
      Update $\theta_d$ by gradient ascent with learning rate $l_2$: minimize $O_g$ \;
}
Update $\theta_e$ and $\theta_c$ by gradient ascent with learning rate $l_3$: 
       minimize $O_a$ \;
} 
\rev{*Note: $n$ and $k$ are two important hyper-parameters}
 \caption{Mini-batch stochastic training of reconstructive adversarial network (RAN)}
 \label{alg_ran}
\end{algorithm}

Our goal with RAN is to train an Encoder that can produce output that is both useful, \ie leading to high inference accuracy when used for classification tasks, and private, \ie leading to high reconstructive error when reverse engineered by an attacker. 
As we noted in \S\ref{sec:problem}, these two objectives can be competing when taken naively.
The key idea of RAN's training algorithm is to train the Encoder along with the Classifier and the Decoder, which simulate the service provider and a malicious attacker, respectively. Given a training dataset $\mathbf{T}$ of $m$ pairs of I, the raw data, and Y, the true label, we train a RAN through an iterative process with three stages:
\begin{enumerate}

\item Discriminative training maximizes the accuracy in Classifier; mathematically, it minimizes the cross entropy $\mathop{H}$ between predicted class $\mathop{C}(\mathop{E}(I_i))$ and true label $Y_i$:
\begin{equation}
\ O_d=\sum_{i=1}^{m} {\mathop{H}(Y_i-\mathop{C}(\mathop{E}(I_i))) }.
\end{equation}

\item Generative training minimizes the reconstructive error by the Decoder:
\begin{equation}
\ O_g=\sum_{i=1}^{m} {|I_i- \mathop{D}(\mathop{E}(I_i)) |^2 }
\end{equation}

\item Adversarial training finds a tradeoff point between utility and privacy: 
\begin{equation}
\ O_a=\sum_{i=1}^{m} {\lambda \mathop{H}|Y_i-\mathop{C}(\mathop{E}(I_i))| - (1-\lambda) |I_i- \mathop{D}(\mathop{E}(I_i))|^2 }
\end{equation}
It is essentially a Lagrangian function of the objectives of the first two stages. $\lambda$ is the Lagrange multiplier that can be used to balance between utility and privacy. 
\end{enumerate}

Algorithm \ref{alg_ran} summarizes the three-stage training algorithm. 
And we leverage mini-batch techniques to 
balance the training robustness and efficiency (line 3)~\cite{bib:li2014:sigkdd}.
Within each epoch, we first perform the discrminative and generative stages (line 5, 6) to initialize model weights.
And then, we perform the adversarial stage (line 8) to seek a balance between utility and privacy.
We note that $k$ in line 4 is a hyper-parameter of first two stages. These $k$ steps followed by a single iteration of the third stage is trying to synchronize the convergence speed of these three training stages well, borrowing existing techniques in generative adversarial network~\cite{bib:goodfellow2014:advances}.
Our implementation uses an overall optimized value, $k=3$, \rev{through comparing several discrete options}.
And we leverage the AdamOptimizer~\cite{bib:kingma2014:arxiv} with an adaptive learning rate for all three stages (line 5, 6 and 8).

\section{Evaluation}
\label{sec:experiment}
%
In this section, we first compare \systemnameposs performance on privacy-utility tradeoff with three baselines and
then visualize the utility and privacy of resulting Encoder output.

\textbf{Evaluation tasks and models.}
We evaluate RAN, especially the resulting Encoder, with five popular classification services.
Specifically, \systemname is evaluated for hand-written digit recognition ($T_{1}$: MNIST ~\cite{data:mnist1998:LeCun}), image classification ($T_{2}$: CIFAR-10~\cite{data:cifar}, $T_{3}$: ImageNet~\cite{data:imagenet}), acoustic event sensing ($T_{4}$: UbiSound~\cite{bib:sicong2017:IMWUT}), and the accelerometer and gyroscope data based human activity recognition ($T_{5}$: Har~\cite{data:Har}). 
\rev{According to the sample size, the LeNet is selected as the neural architectures of RAN's Encoder plus Classifier for $T_1$, $T_4$ and $T_5$, while AlexNet and VGG-16 are chosen for $T_2$ and $T_3$, respectively. To assume a powerful adversary that knows the Encoder in the training, the RAN's Decoder exactly mirrors its Encoder for each task in the training.}

\subsection{Utility vs. Privacy Tradeoffs}
This experiment illustrates the superiority of \systemname compared with three state-of-the-art data privacy preserving baselines.
It does so with five tasks ($T_1, T_2, T_3, T_4, T_5$). However, due to space limit we do not show the results for CIFAR-10 because they are similar to those for ImageNet.
%

\begin{itemize}[leftmargin=*]

\item \textbf{Noisy Data (Noisy)} method \rev{perturbs the raw data, through adding random Laplace noise to the raw data $I$ and then submit the noisy data $\overline{I}$ to the service provider. This is a typical local differential privacy method~\cite{bib:he2017:ACC, bib:dwork2010:STC}.
%
The utility of noisy data is the inference accuracy in a standard deep model,
%
and its privacy is evaluated by the information loss metric, \ie $|I-\overline{I}|^2$. 
%
}


\item \textbf{DNN} method \rev{encodes the raw data into deep features, using a DNN based encoder (\eg the convolutional and pooling layers of LeNet, AlexNet, VGG), and only deliver deep features to the service provider~\cite{url:googlecloud, url:googlenow}. Its privacy is tested by the reconstruction error in a Deconvolutional model (mirrors of the encoder), and the the accuracy evaluates the utility in a DNN based classifier (\eg the fully-connected layers of LeNet, AlexNet, VGG).}


\item \textbf{DNN(resized)} method 
further perturbs above deep features through principal components analysis and Laplace noise injection, and then deliver the perturbed deep features to the service provider~\cite{bib:ossia17:arxiv}. 
Its privacy and utility are also tested by the deep model based decoder and classifier, same with that in the DNN baseline.  

\item \textbf{RAN} \rev{automatically transform the raw data into features, \ie Encoder output, and then deliver them to the service provider. 
The privacy of RAN's Encoder output is tested by the reconstruction error in a separately trained decoder, which is taught based on the binary version (input and output) of the trained RAN's Encoder, to simulate a malicious attacker. 
And its utility is tested by the inference accuracy in \systemnameposs Classifier.}

\end{itemize}
\rev{
The DNN method provides a high utility standard, and the Noisy and DNN(resized) methods set a strict benchmark for RAN.
}


\begin{figure*}[t]
\label{fig_compare_baseline}
  \centering
  \subfloat[Digit(MNIST)]{
  \includegraphics[width=0.48\textwidth]{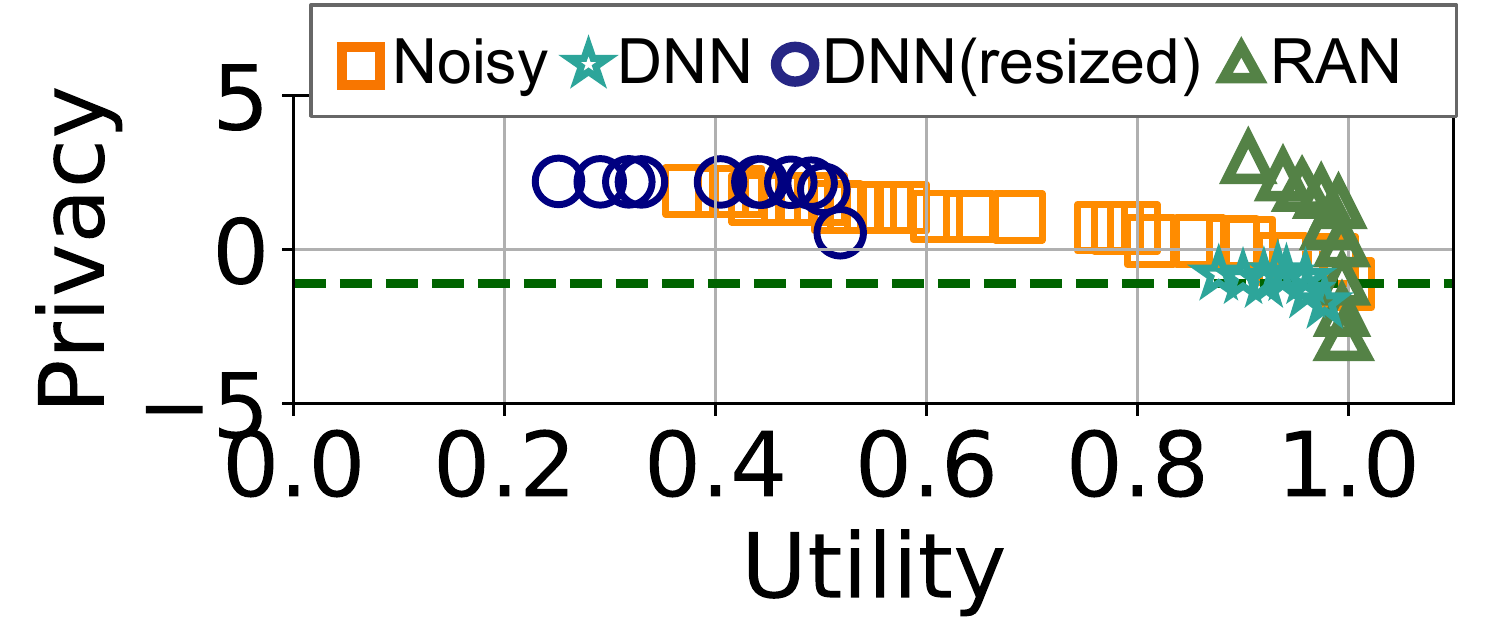}}
    \subfloat[Image(ImageNet)]{
  \includegraphics[width=0.48\textwidth]{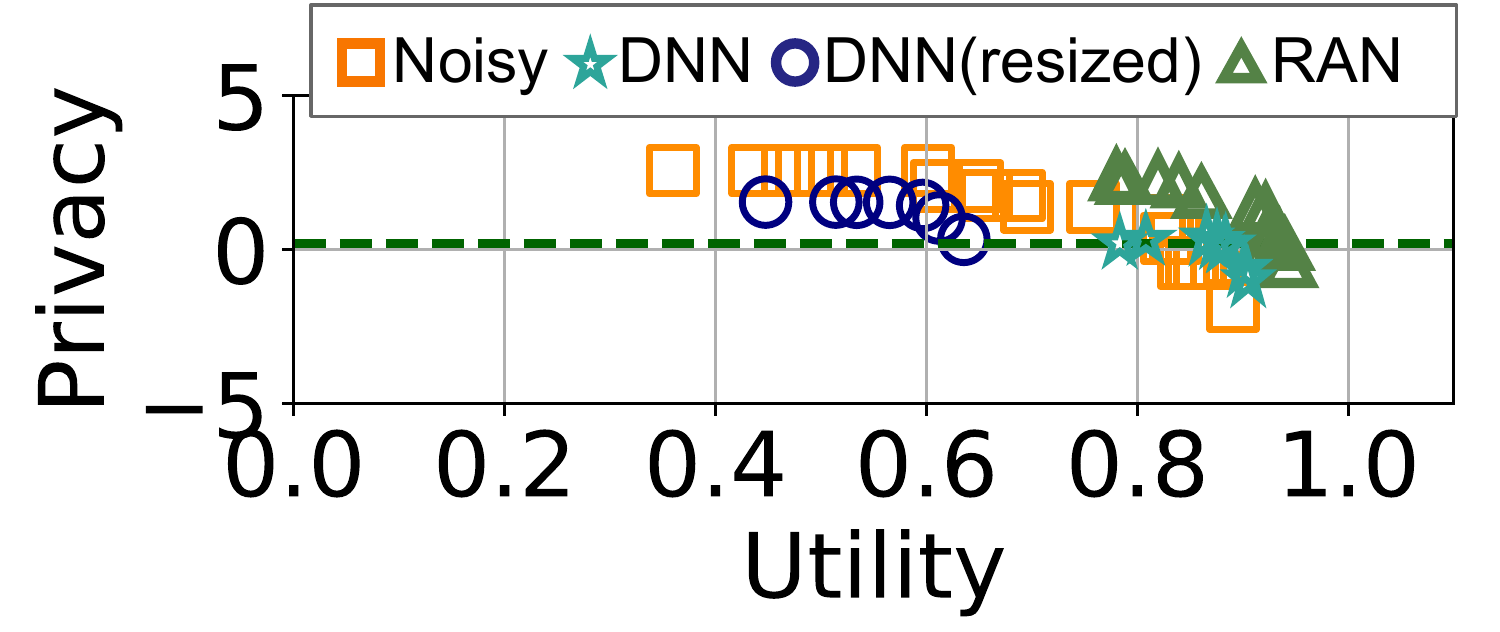}}
   \hfill
    \subfloat[Sound(UbiSound)]{
  \includegraphics[width=0.48\textwidth]{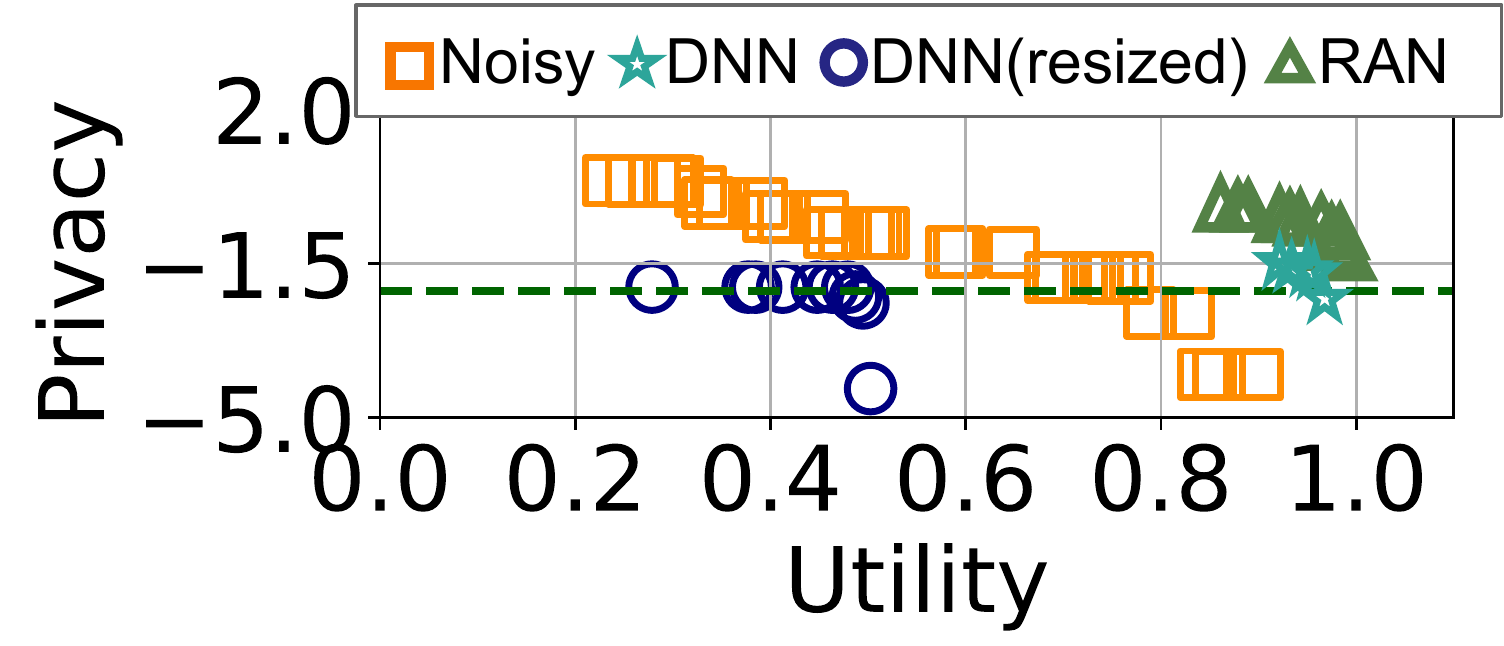}}
    \subfloat[Activity (Har)]{
  \includegraphics[width=0.48\textwidth]{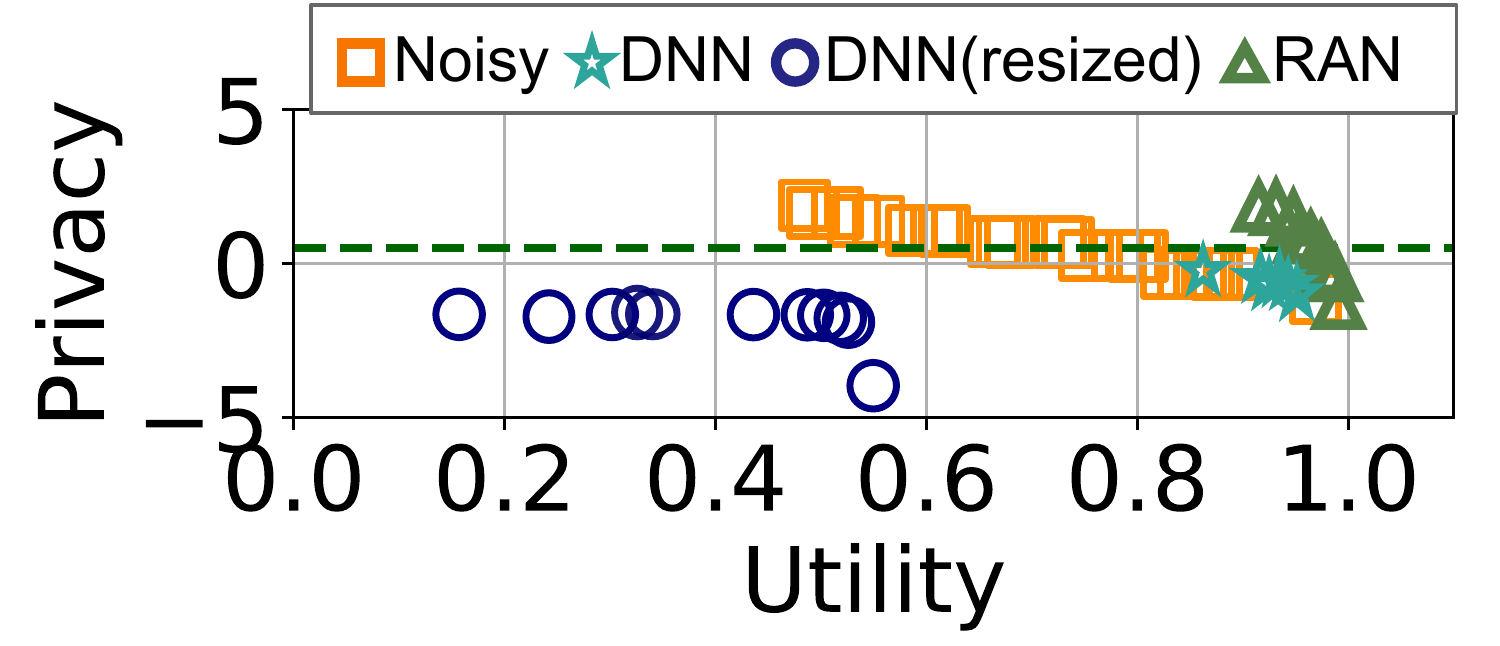}}
\caption{Performance comparison of RAN with three baselines on four datasets (MNIST, ImageNet, UbiSound and Har).
Y-axis is the test reconstruction error, normalized by $log$ operation. And X-axis represents the utility (accuracy).} 
\end{figure*}



\rev{
Figure 3 summarises the Pareto front of the testing privacy-utility tradeoff by using three baselines and \systemname.
In this thread of experiments, we inject various noise factor $\{0.1, 0.2, ... 0.9\}$ into each piece of testing data to test the trained Noisy and DNN(resized) baselines, which are both noise related methods.
And we test RAN models which are trained under several settings $\{ 0.01, 0.02,..., 0.9 \}$ of the Lagrange multiplier $\lambda$ in Eq.4, to recover its tradeoff trends.
First, we see \systemnameposs Encoder output achieves the most stable privacy-utility tradeoff with a constrictive range, compared to those encoded by other three baselines.
Second, \systemnameposs Encoder output achieves the best overall utility than other three baselines. Specifically, RAN's output privacy (utility) is $\geq 95\%$ on MNIST, Ubisound and Har, $\geq 85\%$ on ImageNet, and $\geq 76\%$ on CIFAR-10, with the proper $\lambda$ setups, which is even larger than that of the original deep model (see DNN baseline).
While the accuracy in Noisy and DNN(resized) baselines is unstable, ranging from $20\%$ to $93\%$.
Third, \systemnameposs output can attain the higher privacy than usual deep features in a traditional DNN, and guarantee competitive privacy compared to others. 
Moreover, the RAN's privacy quantified by RAN's Decoder (the green dashed line in Figure 3) is averagely larger than that measured by a third-party Decoder (green triangles in Figure 3) which is trained given the binary version of RAN's Encoder.
}

%

\textbf{Summary.} \rev{ Overall, \systemname outperforms other three baselines to attain a better privacy-utility tradeoff over five recognition tasks. 
Second, the features derived by the proposed learning algorithm to remove redundant (sensitive) information, for privacy, even surpass the accuracy of the original model. We refer readers to $\S$~\ref{sec:discuss} for how and why it works from a theoretical perspective.
We also note that the regularization parameter $\lambda$ in RAN can be further systematically fine-tuned, e.g., exponentially varied using reinforcement learning, so that discovers a better privacy-utility tradeoff.
}
%


\begin{figure*}[t]
  \centering
  \includegraphics[width=0.88\textwidth]{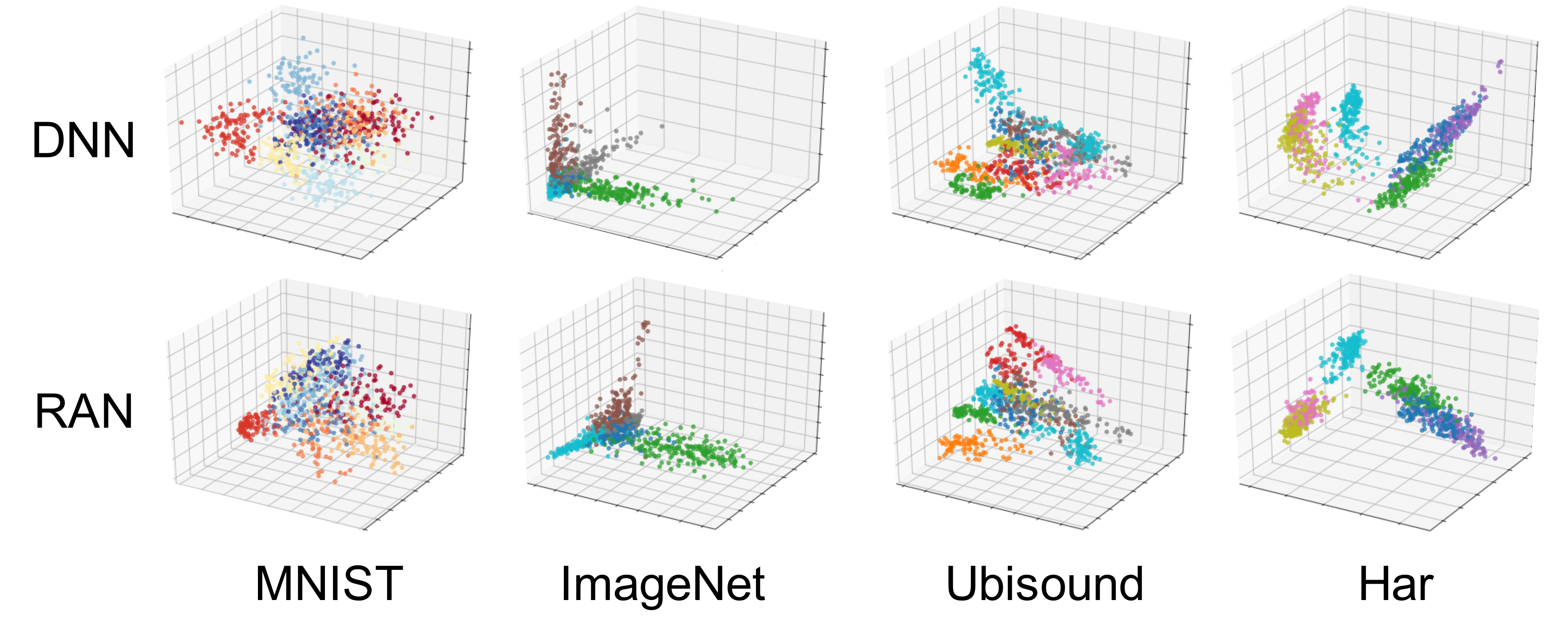}
  \caption{3D visualization of the highly separable features learned by standard DNN and RAN's Encoder output in the feature space. Different color in each figure standards for one class.}
\label{fig_diff_feature}
\end{figure*}

\begin{figure*}[t]
  \centering
  \includegraphics[width=0.95\textwidth]{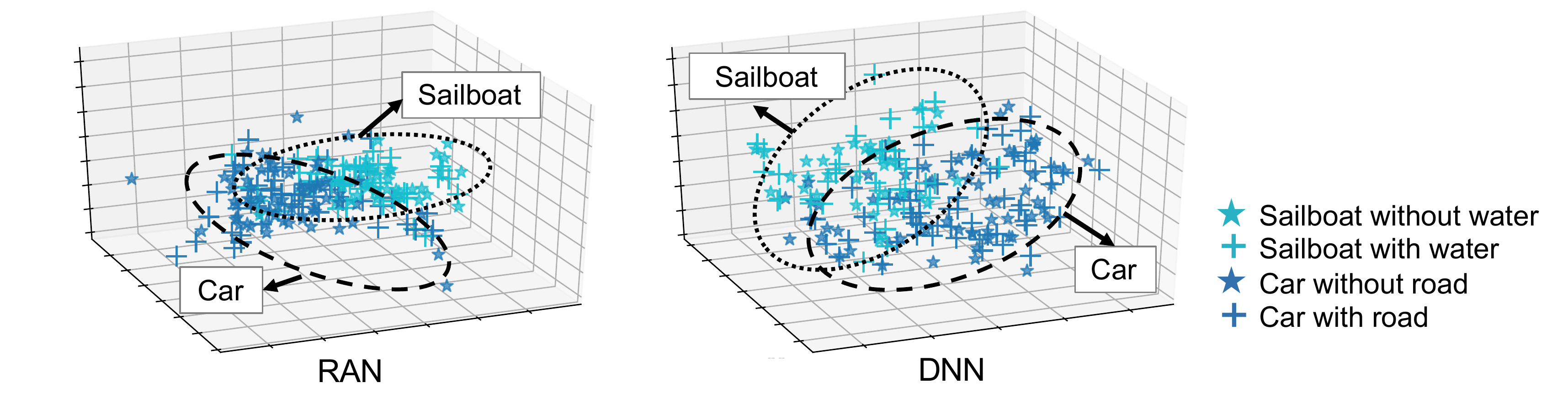}
  \caption{Zoom in on two categories, \ie sailboat and car in the feature space.}
\label{fig_boat_feature}
\end{figure*}

\begin{figure*}[t]
  \centering
  \includegraphics[width=0.95\textwidth]{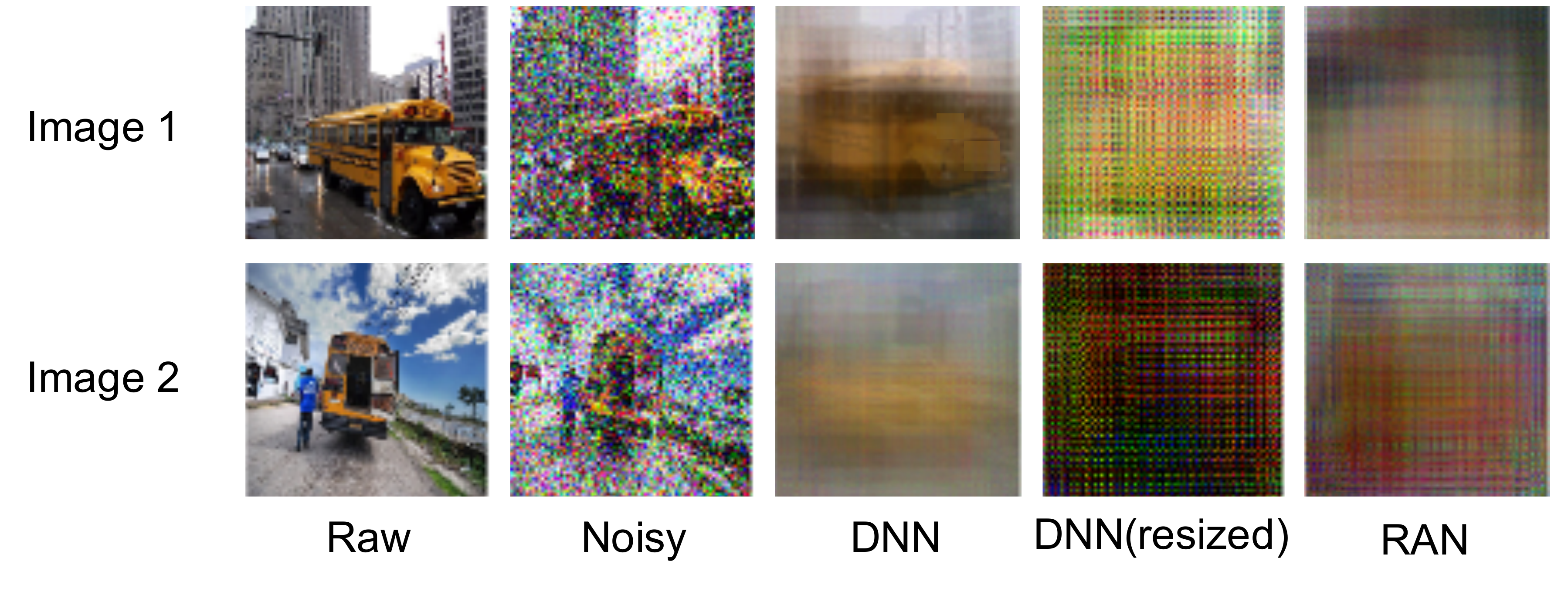}
  \caption{From left to right: raw image from ImageNet (Raw), image with Laplace noise (Noisy), images reconstructed from DNN's features, resized DNN's features, and RAN's Encoder output. }
\label{fig_diff_pixel}
\end{figure*}

\subsection{Utility Visualization of RAN's Encoder output}
\label{sec:vis_utility}
\rev{
To illustrate the utility of \systemnameposs Encoder output, we visualize how the distribution of RAN's Encoder output varies from  traditional Depp features.
First, as shown in Figure~\ref{fig_diff_feature}, \systemnameposs Encoder output are highly separable, in the feature space, similar to the deep features from traditional DNN. It reflects the utility for subsequent classification.
Second, to zoom in on two categories of images for more details in Figure~\ref{fig_boat_feature}, we see that RAN pushes the features towards the constrictive space dominant dominated by the data without redundant information, \ie "sailboat without water" and "car without road".
While the traditional DNN may capture the background "water" and "road" information to help the classification of "sailboat" and "car".
}
%
%
%

\textbf{Summary}. \rev{First, \systemnameposs Encoder output is highly separable in feature space as standard DNN do, which indicates the high utility for the subsequent classification tasks.
Second, the learning algorithm on RAN pushes features towards essential information and away from redundant background (sensitive) information (see more interpretations in $\S$~ \ref{sec:discuss}).}

\subsection{Visualizing of RAN's Privacy}
\label{sec:vis_pivacy}
%
In this experiment, we visualize the privacy of \systemnameposs Encoder output, i.e. private features, in comparison to other approaches, using
two example images from ImageNet. 
Figure~\ref{fig_diff_pixel} illustrates the pixel image of the raw data, the noisy data, the mimic data reconstructed from DNN's deep feature, and mimic data reconstructed from RAN's private features from two "bus" images from ImageNet datasets.
%
%
We can find that the image reconstructed by \systemnameposs Decoder are dramatically corrupted and hard to distinguish the exact information of raw images.
%
As mentioned in $\S$ 3.1, the \systemnameposs Decoder is more potent than a separately trained Decoder on reconstructing RAN's hidden features.
%
%

\textbf{Summary}. First, the corrupted reconstructed images by \systemname certify the improved privacy of \systemnameposs Encoder output.
Second, the reconstructed images from DNN's features recover both object (bus) and background (road) information, while RAN's Encoder tries to contain object information and remove background information. And then RAN leads better privacy and utility (generalization) to its hidden features. More interpretation is in  $\S$~\ref{sec:discuss}.
%
\section{Manifold Based Interpretation}
\label{sec:discuss}

%

We resort to the manifold perspective of the deep model. It is common in literature to assume that the high-dimensional raw data lies on a lower dimensional manifold, refers to latent variables~\cite{bib:chien2016:ICASSP}.
A DNN can also be viewed as a parametric manifold learner utilizing the nonlinear mapping with multi-layer architecture and connection weights.
%
%

%
We decompose the input data $I$ into two orthogonal lower dimensional manifolds:
\begin{equation}
I =  I^{OD} +  I^{OD}_{\perp}
\end{equation}
Here, the component $I^{OD}$ is the ideal manifold component that is both necessary and sufficient for object detection. Thus, ideally, we want our training algorithm to rely on this information for object detection solely. Formally, for the discriminative classifier, this implies that $prob(Y|I) = prob(Y|I^{OD})$. 
%
And the other manifold component $I^{OD}_{\perp}$,
orthogonal to $I^{OD}$, may or may not contain information for the object class, but it is dispensable for object detection. 
In practice, the real data does have redundant correlations. Thus $I^{OD}_{\perp}$ may be learned for object detection, but unnecessary. 
However, revealing $I^{OD}_{\perp}$ is likely to contain sensitive information thus hurt the privacy.
If we assume that there does exist a sweet-spot trade-off between utility and privacy, that we hope to find, then it must be the case that $I^{OD}$ is not sensitive (as it is necessary and sufficient). 

The features $F$ learned by standard deep learning algorithms to minimize the training error based on information from $I$, will mostly likely overlap (non-zero projection) with both $I^{OD}$ and $I^{OD}_{\perp}$. And the overlap with $I^{OD}_{\perp}$ compromises the privacy (as evident from our experiments).  
Apart from privacy, the redundant correlation in $I^{OD}_{\perp}$ is also likely only be spurious in training data. Thus, merely minimizing training loss can lead to over-fitting.

%

\begin{figure}[th]
\centering
\includegraphics[width=.98\textwidth]{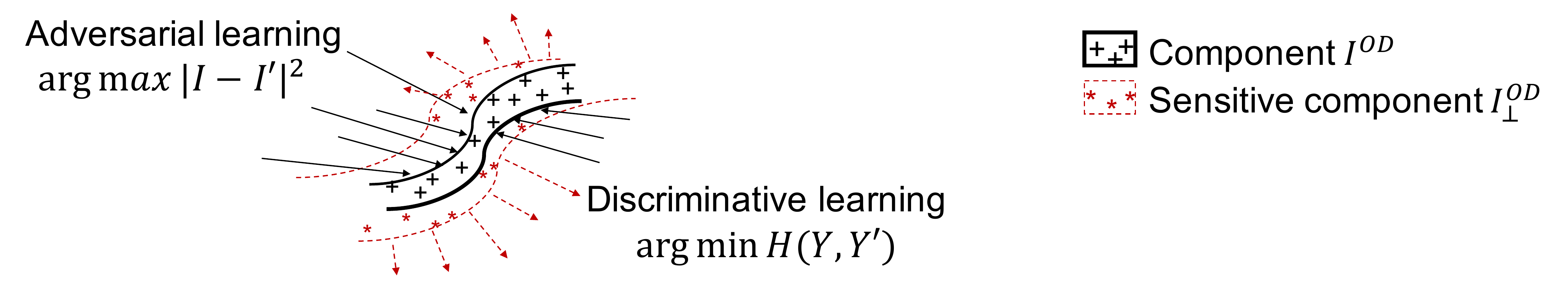}
\caption{A new manifold pushed by \systemname to form the feature extractor, \ie Encoder, for utility and privacy.
The utility-specified discriminative learning objective (Eq.(2)) push it to contain $I^{OD}$ and $I^{OD}_{\perp}$,
and the privacy-imposed adversarial training objective (Eq(4)) pushes it away from sensitive component $I^{OD}_{\perp}$.
}
\label{fig_manifold}
\end{figure}

This is where we can shoot two stones via an adversarial process. In \systemname, the Encoder is trained by utility-specified discriminative learning objective (Eq.(2)) and privacy-imposed adversarial learning objective (Eq.(4)), to find features $F'$ as shown in Figure~\ref{fig_manifold}.
%
The manifold $I'$ formulated by parametric Encoder is forced by discriminative learning objective (Eq.(2)), just like the traditional approach, to contain information from both $I^{OD}$ as well as $I^{OD}_{\perp}$. However, the adversarial training objective (Eq(4))
will push features $F'$ away (or orthogonal) from $I^{OD}_{\perp}$. In this way, we get privacy as well, since $F'$ as a function of $I$ which has two manifolds, being orthogonal to $I^{OD}_{\perp}$ forces it to only depend on $I^{OD}$. 

Meanwhile, from a generalization perspective, 
in the training data, 
%
the spurious information from $I^{OD}_{\perp}$ that might over-fit the training data is iteratively removed by the adversarial training objective (Eq.(4)) automatically leading to enhanced generalization.
For example, as shown in Figure~\ref{fig_diff_pixel}, if we want to discriminate between "bus" and "sailboat", the "road" in the picture can help, but it is obviously a bad way of classifying and may not generalize if the test image contains "bus" without the "road". 
However, "road" maybe most of the background and retain some information to ease image reconstruction, which is unintended. Adding noise will obfuscate both "road" and "bus", compromising object detection at the cost of privacy. 
The RAN, instead, will only obfuscate "road", making reconstruction impossible without compromising the utility. In fact, RAN will get increased utility due to better generalization. 

This is an auspicious illustration that in machine learning we can gain both utility and privacy in practice. A rigorous formalism and study of this phenomena could be an independent field in itself.


\section{Related Work}
Our work is closely related to the following categories of research. 


\textbf{Privacy Preserving for Mobile Data}:~~
Unlike the typical privacy preserving techniques which are adopted by data collectors (service providers) to release data for public data mining, \systemname keeps the raw data under end-user's control, \ie the the user submits private features only, rather than raw data, to service providers.
%
%
For example, randomized noise addition~\cite{bib:he2017:ACC} and Differential privacy~\cite{bib:dwork2014:FTTCS,bib:abadi2016:abadi} techniques have been widely used by service providers to anonymize/remove personally identifiable information or only releases statistical information to publicly release datasets. 
\systemname outperforms Noisy data (a differential privacy method) with better classification utility and competitive privacy ($\S$ 3.1), because \systemnameposs Encoder is end-to-end trained with collaborative utility-specified deep learning and privacy-imposed adversarial learning for a good trade-off between features’ utility and privacy.

\textbf{Privacy Preserving with Deep Learning}:~~
%
%
Generally, 
prior works adopt two classes of approaches to protect end-user's raw data: the end user modifies raw data before delivering them to service providers~\cite{bib:ossia17:arxiv} or multiple end users cooperate to learn a global data mining results, without revealing their individual raw data~\cite{bib:li2017:FGCS}.
However, these segmented systematic methods inevitably incur utility drops in subsequent recognition tasks.
We has compared RAN with resized noisy deep features according to~\cite{bib:ossia17:arxiv} (\S 3.1), and concluded \systemname achieves a better utility against altering raw data into resized deep features.
This is because \systemnameposs Encoder is also trained along with a accuracy discriminator (Classifier) to guarantee utility.

%
%
%
%
%
%
%

\textbf{Deep Feature Learning Techniques}:~~
%
In order to generate special features to facilitate the subsequent classification utility and protect raw data's sensitive information from recovering by generative models,
\systemname is the first to present an end-to-end deep architecture to sidestep the black-box of collaborative discriminative and generative learning via an end-to-end adversarial process. 
Today's extensions of discriminative models, generative models, or both, have been studied to seek latent feature variables,
which contributes to inference accuracy but incurs easy data reconstruction by reverse techniques~\cite{bib:radford2015:arxiv, bib:zhong2016:FDS}.
%
%
And some components used in existing generative models, such as sensitivity penalty in contractive autoencoder~\cite{bib:rifai2011:ICML}, data probability distribution in generative adversarial network~\cite{bib:goodfellow2014:advances} and KL divergence in variational autoencoder~\cite{bib:doersch2016:arxiv}, can be further integrated into \systemnameposs framework to define and enhance application-based privacy.

%
%
%

%


\section{Conclusion}
This paper presents to establish a deep model for mobile data contributor, \ie mobile users, to encode the raw data into perturbed features before delivering it to the data collector or miner, \ie service provider.
To realize it, we present \systemname, a novel deep model for private and useful feature transforming.
\systemname is the first to not only maximize feature's classification accuracy but also maximize its reconstruction error via an end-to-end adversarial training process.
In particular, \systemname consists an Encoder for feature extracting, a Decoder for data reconstruction error (privacy) quantification from Encoder output and a Classifier for accuracy (utility) discrimination.
The proposed training algorithm upon \systemnameposs contains three phase: discriminative learning function on Encoder and Classifier to boost their discriminative abilities, a generative stage on Decoder to improve its data generative capacity which stand in the position of Encoder's adversary, and an adversarial stage on Encoder, Classifier and Decoder to achieve our design objectives.
%
%
Evaluations on five widely used datasets show that \systemnameposs Encoder output attains a notable privacy-utility tradeoff.
In the future, we plan to investigate finer-grained manifold learning techniques on \systemname for feature generalization and privacy improvements.

A few aspects of \systemname do invite further research.
%
%
%
%
\rev{First, the RAN framework and the training algorithm can accommodate different choices of privacy quantification, especially application-specific ones. For example, we could measure the privacy by the hidden failure, \ie the ratio between the background patterns that were discovered based on RAN’s Encoder output, and the sensitive patterns founded from the raw data, in an object recognition task.}
Second, the training of two adversaries in \systemnameposs, \ie Encoder and Decoder, must be synchronized well to avoid model degradation.
%
It is because of the convergence diversity of Encoder and Decoder. Therefore, some more efforts are needed in \systemname, \eg setting proper iteration steps $k$ and learning rate.

%
%
%
%


\bibliography{iclr2019_conference}
\bibliographystyle{iclr2019_conference}


\end{document}